\documentclass[letterpaper]{article} 
\usepackage{aaai24}  
\usepackage{times}  
\usepackage{helvet}  
\usepackage{courier}  
\usepackage[hyphens]{url}  
\usepackage{graphicx} 
\usepackage{natbib}  
\usepackage{caption} 
\usepackage{algorithm}
\usepackage{algorithmic}
\usepackage{dblfloatfix}
\usepackage{newfloat}
\usepackage{listings}
\DeclareCaptionStyle{ruled}{labelfont=normalfont,labelsep=colon,strut=off} 
\floatstyle{ruled}
\newfloat{listing}{tb}{lst}{}
\floatname{listing}{Listing}
\usepackage{multirow}
\usepackage{booktabs}
\usepackage{amssymb}
\usepackage{pifont}
\newcommand{\xmark}{\ding{55}}%

\title{Using Artificial Populations to Study Psychological Phenomena in Neural Models}
\author {
    Jesse Roberts\textsuperscript{\rm 1},
    Kyle Moore\textsuperscript{\rm 1},
    Drew Wilenzick\textsuperscript{\rm 2},
    Doug Fisher\textsuperscript{\rm 1}
}
\affiliations {
    \textsuperscript{\rm 1}Vanderbilt University\\
    \textsuperscript{\rm 2}Cornell University\\
    jesse.roberts@vanderbilt.edu, ddw77@cornell.edu, kyle.a.moore@vanderbilt.edu, douglas.h.fisher@vanderbilt.edu
}

\usepackage[switch]{lineno} 

\nocopyright
\begin{document}
\maketitle
\begin{abstract}
The recent proliferation of research into transformer based natural language processing has led to a number of studies which attempt to detect the presence of human-like cognitive behavior in the models. We contend that, as is true of human psychology, the investigation of cognitive behavior in language models must be conducted in an appropriate population of an appropriate size for the results to be meaningful. We leverage work in uncertainty estimation in a novel approach to efficiently construct experimental populations. The resultant tool, PopulationLM, has been made open source. We provide theoretical grounding in the uncertainty estimation literature and motivation from current cognitive work regarding language models. We discuss the methodological lessons from other scientific communities and attempt to demonstrate their application to two artificial population studies. Through population based experimentation we find that language models exhibit behavior consistent with typicality effects among categories highly represented in training. However, we find that language models don't tend to exhibit structural priming effects. Generally, our results show that single models tend to over estimate the presence of cognitive behaviors in neural models. 
\end{abstract}

\section{Introduction}

In the wake of success following the introduction of transformers in \cite{vaswani2017attention} and the public deployment of powerful variants of GPT, many have started to question if these models exhibit behavior similar to human cognition.

Work analyzing the cognition of these powerful models is important not only to the explanation and interpretation of the models themselves but may offer insight into cognition more generally, a synergism best embodied by the interplay of reinforcement learning and neuroscience \cite{subramanian2022reinforcement}. We believe this emerging study of cognitive behavior in neural models can be improved by adopting methods from branches of science more typically associated with statistical testing. Without appropriate experimental methodology conclusions may not be robust in the face of variations, a symptom associated with the greater replicability crisis \cite{goodman2016does}. Research attempting replication and extension of ToM results in GPT-4 found that relatively small experimental alterations caused the effect to disappear \cite{ullman2023large}. This suggests the experimental design in the original study was insufficient to support the drawn conclusions. 

This is precisely the motivation for the present paper. Claims that fail to be reproducible regarding powerful AI models may ultimately result in erosion of the public's trust and attention. Any study of neural model cognitive behavior should characterize not only the presence but the size of the effect and the significance. Doing so necessitates rigor which may decrease erroneous conclusions, and will permit better explanation of neural model cognitive behavior through meta-analytic study. To this end, we present and demonstrate an artificial population generation method for the study of cognitive phenomena in neural models with the hope that it will aid in the reproducibility of research regarding the behavior of neural models.

This paper contributes by drawing connections between social and behavioral experimental design and neural model uncertainty estimation resulting in a (1) tool called PopulationLM for the creation of populations of neural models via stratified MC dropout. We harvest novel metrics and explore population best practices by applying artificial populations to the (2) replication and extension of \cite{misra2021language} (correlation analysis) and (3) \cite{sinclair2022structural} (difference analysis). We present novel results regarding the presence of typicality and structural priming effects in language models. 


\section{Behavioral Phenomena in Neural Models}

\begin{table*}[t]
    \centering
    \begin{tabular}{l|c|c|c|c|c}
        Phenomena &
        Study by &
        \begin{tabular}[c]{@{}c@{}}Measure(s)\end{tabular} &
        \begin{tabular}[c]{@{}c@{}}Statistic \end{tabular} &
        \begin{tabular}[c]{@{}c@{}}Significance \end{tabular} &
        \begin{tabular}[c]{@{}c@{}}Experimental Var \end{tabular} \\
        \toprule
        \multirow{5}{*}{Theory of Mind}
            & \citeauthor{bubeck2023sparks}       & qualitative             & ---                & ---      & not isolated   \\
            & \citeauthor{kosinski2023theory}     & frequency             & ---                & ---      & not isolated   \\
            & \citeauthor{sap2022neural}          & frequency               & ---                & ---      & not isolated   \\
            & \citeauthor{ullman2023large}        & frequency               & ---                & ---      & isolated*      \\
            & \citeauthor{trott2023large}         & token probs             & $\chi^2$ + $\beta$ & reported & not isolated   \\ 
        \hline 
        \multirow{3}{*}{\begin{tabular}[c]{@{}l@{}}Logical  Reasoning\end{tabular}}
            & \citeauthor{binz2023using}          & token probs             & $\chi^2$ + $t$ + $\beta$ & reported & isolated*      \\
            & \citeauthor{mccoy2019right}         & frequency               & ---       & ---   & isolated       \\
            & \citeauthor{lamprinidis2023llm}     & frequency               & ---       & ---   & not isolated   \\
        \hline
        \multirow{3}{*}{\begin{tabular}[c]{@{}l@{}}Framing \& \\ Anchoring\end{tabular}}
            & \citeauthor{binz2023using}          & token probs            & $\chi^2$ + $t$ + $\beta$ & reported       & isolated*      \\
            & \citeauthor{jones2022capturing}     & frequency             & ---       & ---   & isolated       \\
            & \citeauthor{suri2023large}          & frequency             & $t$         & reported       & isolated*      \\
        \hline
        \multirow{2}{*}{\begin{tabular}[c]{@{}l@{}}Decision-making\\ Heuristics\end{tabular}}
            & \citeauthor{binz2023using}          & token probs             & $\chi^2$ + $t$ + $\beta$ & reported       & isolated*      \\
            & \citeauthor{jones2022capturing}     & frequency             & ---       & ---   & isolated       \\
        \hline
        Typicality
            & \citeauthor{misra2021language}      & token probs             & $r$ + $\rho$  & reported       & isolated       \\
        \hline
        Priming
            & \citeauthor{sinclair2022structural} & token probs             & ---       & ---   & isolated       \\
        \hline
        \begin{tabular}[c]{@{}l@{}}Emotion  Induction\end{tabular}
            & \citeauthor{coda2023inducing}       & frequency             & $r$ + $t$ + probit $\beta$   & reported      & not isolated   \\
    \end{tabular}
    \caption{Review summary of large language model behavioral studies. $r$ = Pearson, $\rho$ = Spearman, $\beta$ = Berksons, $t$ = t-test.}
    \label{tab:behav-rev}
\end{table*}

In this section we review the current work related to the study of cognitive behavior in neural language models paying specific attention to the measures reported and methodology. Table \ref{tab:behav-rev} summarizes the works that have been identified, organized by the behavioral phenomenon that they investigate. This review and meta-commentary does not invalidate any of the findings in the associated papers. Rather, it serves as a compendium of work so far in this field and helps to illuminate the problem we wish to address. 

The measures reported refers to the measure applied to the model output. Statistic refers to the statistical analysis applied to the measures. We find that most papers used atypical measures of effect like frequency of occurrence or qualitative analysis and tend to not use statistical testing. Those employing t-tests don't typically specify the particular test. Analogously, we find that less than a third of the papers report significance levels for their results. In contrast, most authors did isolate the experimental, independent variable. Rows marked with an * indicate works that did so only in a subset of reported experiments. 

No study in our review utilizes uncertainty estimation to systematically perturb the model or the input. Therefore, no work has been done to study neural behavior in a population. In the latter half of this paper, we study two behavioral phenomena from table \ref{tab:behav-rev} in artificial populations: typicality and structural priming (SP). Typicality refers to a high degree of agreement across subjects in humans when ranking items as more or less typical of a given category and is known to be related to rate of retrieval of an item given the category \cite{roschCognitive1975}. Structural priming refers to the predilection for a sentence structure similar to the most recently observed syntactical structure \cite{pickering2008structural}.

\section{Populations of Neural Models}

In all social and behavioral science, conclusions drawn from a single subject face severe limitations. Without a population of subjects it is impossible to know if the individual is typical along the dependent variable in the population or an outlier. 

Studying the cognitive behavior of neural models, either as an ontology or in relation to human psychology, suffers from a similar limitation. There always exists a possibility that an expression of a behavior is anomalous or that the behavior is tenuously supported in the network. 

In this paper we refer to models and their derivatives as different species. i.e. BERT and DistilBERT are individual species, while they both belong to the same family. Genus is reserved for fine-tuned variants.

Forming inter-species populations is an intuitive but flawed approach since we wish to facilitate the study behaviors that may emerge in specific species, as is known to occur as a function of model size \cite{wei2022emergent}. Inter-species populations don't permit this type of myopic study. 

Instead we form populations using work from neural model uncertainty estimation. In that context, the goal is not a population but an estimation of model uncertainty. However, this is precisely the characteristic typically extracted from a population study, the degree to which a result is consistent across individuals. We refer to this as the population uncertainty. Several uncertainty estimation methods have been proposed in literature and can be placed into 4 broad groups \cite{gawlikowskiSurveyDeep2023}, single network deterministic, test time augmentation, ensemble techniques, and Bayesian approximations. 

Single network deterministic methods attempt to estimate the uncertainty of a network without multiple predictions being made. However, they trade accuracy for speed. Test time augmentation methods perform perturbations of the input data and estimate uncertainty across a single model's outputs \cite{lyzhov2020greedy}. Though this is a promising solution for closed source models, there exists a bound on the perturbation resolution possible in transformers with test time augmentation due to Hahn's lemma \cite{hahnTheoretical2020}. Ensemble techniques, generally outperform Bayesian methods \cite{lakshminarayanan2017simple} but require multiple models trained independently. The price associated with from scratch training makes this a poor solution \cite{sharir2020cost}. Therefore, Bayesian approximation is the most applicable uncertainty estimation technique for the creation of populations of open source models.

\begin{table*}[t]
\centering
\begin{tabular}{r|c|c|c|c|c}
    Model Species & Paper & Typicality KS test & SP KS test & Type (parameters) & Training Data\\
    \toprule
    DistilBERT & \citeauthor{sanh2019distilbert}  & 0.056 (p$\approx$0.055) \xmark & 0.04 (p$<$0.05) & MLM (66M) & \multirow{3}{*}{BookCorpus, Wiki} \\
    BERT Base & \multirow{2}{*}{\citeauthor{devlin2018bert}} & 0.051 (p$\approx$0.108) \xmark & 0.03 (p$\approx$0.06) \xmark & MLM (110M) & \\
    BERT Large & & 0.072 (p$<$0.01) & 0.05 (p$<$0.01) & MLM (340M) &   \\
    
    \hline
    GPT & \citeauthor{radford2018improving} & 0.069 (p$<$0.01) & 0.08 (p$<$0.01) & MLM (120M) & BookCorpus  \\
    
    \hline
    DistilGPT-2 & \citeauthor{sanh2019distilbert} & -0.072 (p$<$0.01) & 0.45 (p$<$0.01)  & CLM (82M) & \multirow{3}{*}{BookCorpus, WebText}  \\
    GPT-2 & \multirow{2}{*}{\citeauthor{radford2019language}} & -0.03 (p$\approx$0.685) \xmark & 0.29 (p$\approx$0.1) \xmark & CLM (117M) &  \\
    GPT-2 Medium & & 0.075 (p$<$0.01) & \textbf{0.51 (p$<$0.01)} & CLM (345M) &  \\
    
    \hline
    RoBERTa Base & \multirow{2}{*}{\citeauthor{liu2019roberta}} & 0.065 (p$<$0.02) & 0.08 (p$<$0.01)& MLM (125M) & \multirow{2}{*}{\begin{tabular}[c]{@{}c@{}}BERT train data, Stories,\\ CC, OpenWebText, News\end{tabular}} \\
    
    RoBERTa Large & & \textbf{0.15 (p$<$0.1)} & 0.19 (p$<$0.1) & MLM (355M) &  \\
\end{tabular}
\caption{Kolmogorov-Smirnov test for each population and each experiment compared to the base model. Null hypothesis $H_0$ is population probabilities and base model probabilities are drawn from the same underlying distribution per species. Populations very similar to the base model have an \xmark.}
\label{tab:modeldif}
\end{table*}

\section{Population Dropout}

We use Monte Carlo (MC) dropout \cite{gal2016dropout}  to form populations from base models. A neuron mask is assembled from the instances of random variables and placed on the network. The resulting masked network is then used to perform inference. Each network mask is typically applied once and discarded. However, in the context of behavioral studies, it is desirable to apply a set of stimuli to the static population for within-group, paired-sample tests. We contribute stratified MC dropout, a variation that generates and maintains a user defined number of masks for any PyTorch compatible network. While the provided library is implemented only for PyTorch, the method is, in principle, applicable to any neural network library that supports inference-time dropout.

While it is true that dropout populations approximate the distribution of a deep Gaussian process \cite{gal2016dropout}, the degree to which this will approximate a group of humans is not known. Therefore, we don't claim that this method approximates results typical of human studies. We claim that evaluating the dropout population outputs as a group will help the results to be more robust in the face of variation due to decreased presence of poorly supported behaviors as a direct consequence of their tendency to converge to a Gaussian process. 

We do not apply any aggregation to the population outputs. Instead, we adopt methodologies from psychological and pharmaceutical domains to treat the model responses as populations of individuals and directly apply statistical analysis. This approach provides a more robust view of expected model behavior under variation with improved insights regarding population certainty and statistical significance.

\subsection{Analysis of the populations}

We evaluated the efficacy of the populations to generate outputs which are statistically distinct from the base models for each species via the non-parametric Kolmogorov–Smirnov (KS) test. It compares the shape and location of two distributions but makes no assumptions about the nature of the underlying distributions. The null hypothesis is that the sample distributions will have similar shape and location.

\begin{figure}[h]
\centering
\includegraphics[width=\columnwidth]{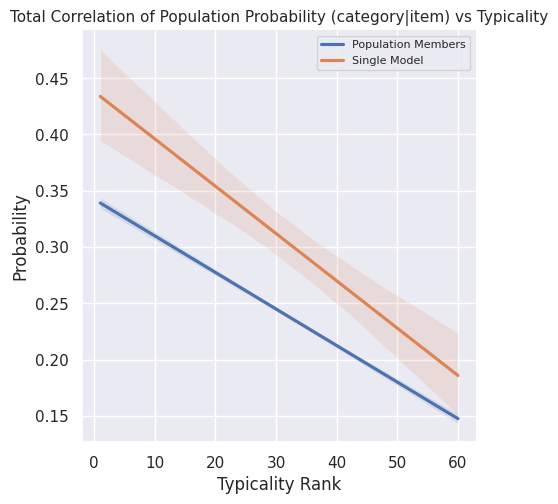} 
\caption{Single model regression vs population of models probability-typicality regression for RoBERTa Large. Rank is inversely related to typicality. 95\% confidence intervals shown for both with very narrow bounds on the population.}
\label{popAnalys}
\end{figure}

In table \ref{tab:modeldif}, we find that the underlying distributions for the species' base models and their populations are not representative of the same distribution with the exception of GPT-2, BERT base, and DistilBERT as judged by the significance of the p value. We inspected these results by observing plots of each and include RoBERTa in figure \ref{popAnalys} juxtaposed with its associated dropout population. An obvious benefit of the population is the narrowing of the confidence bounds on the regression due to augmented elimination of alternative regressions. As suggested by the KS test, the relationship between typicality and population probability in experiment 1 is shown to be quite distinct from that of the base model.

The KS test is a useful method to characterize the likelihood with which population results will vary from the base model on a given task, with a high effect size indicating high likelihood. However, the test is meaningful for the target behavior or context only. This is evinced by the large change in effect among the GPT-2 family from the typicality experiment to the structural priming in table \ref{tab:modeldif}.

\begin{figure*}[t]
\centering
\includegraphics[width=\textwidth]{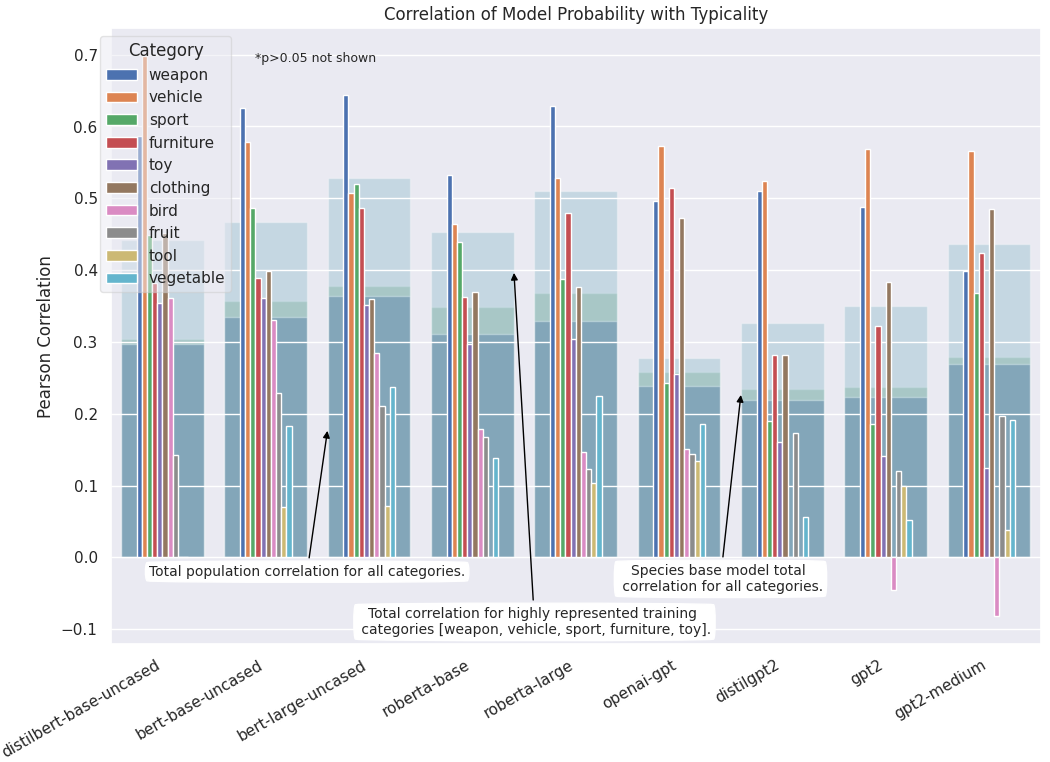} 
\caption{For each model the, colored bars show within category Pearson correlation (p$<$0.03). For each model the total Pearson correlation (p$<$0.03) is shown as the gray background bar. The total Pearson correlation (p$<$0.01) for well understood categories (categories with an average item frequency $>$ 60000 in training data for Bert) is shown as the light blue bar. In well understood categories, typicality of the item may explain up to $r^2 \approx$ 20\% of the category probability volatility.}
\label{E1RQ1}
\end{figure*}

\subsubsection{How much dropout is necessary?}

MC dropout has been applied to transformers previously in \cite{shelmanovHowCertainYour2021} and \cite{vazhentsevUncertainty2022}. In both of these papers the authors experimentally found 0.1 to be the most effective dropout rate for discouraging incorrect, poorly supported outputs. 

We experimented with dropout rates from 0.1 to 0.8. We found no advantage in using larger rates of dropout for experiments, as increased rates caused signal erosion with all behavioral correlations being dissolved beyond rates of 0.5 Therefore, we recommend that statistical studies adopt a 0.1 nominal dropout rate.

\subsubsection{How big should the artificial population be?}

Population size for a study is related to two important statistical measures, significance and power. The significance of a \textit{result} is a measure of the probability of the null hypothesis. The power of a \textit{test} is a measure of the probability that the test will correctly reject the null hypothesis and avoid a false negative. 

For evaluating cognitive behavior in neural models, the power of a test is less important as the effect of a false negative is not likely to cause damage. However, the significance of a result is of the utmost importance as this permits meta-analytical extension and can act as a mitigator of sensationalism when applied properly. 

We empirically find that a population of 50 is an acceptable compromise, providing sufficient statistically significant deviation from the base model in table \ref{tab:modeldif} without dramatic computational costs. Interestingly, it seems the models tend to have correlated relationships with associated dropout populations. The KS tests for the two experiments in table \ref{tab:modeldif} show that BERT, GPT-2, and RoBERTa models tended to have KS effect sizes which were rank correlated across experimental populations within model families. However, the model correlations don't extend outside the family. This suggests that 50 member populations may tend to be sufficient for the approximated deep Gaussian process to emerge.

\section{Experiment 1: Typicality Effects in Language Models}

We reproduce and extend the experiment conducted in \cite{misra2021language} which assessed the base model total correlation between probability and typicality. Our base model probabilities agree with past results, and we contribute novel tests using dropout populations and within category analysis which shed light on the factors that support the emergence of typicality effects in language models.

\subsubsection{Experimental Setup} 

We use typicality data from \cite{roschCognitive1975} which gives a typicality rank, $r_i$, for each item, $i$, in category $C$. As in the original experiment, we construct prompts, $\pi_i$, for each $i\in C$ and measure the probability assigned to the category given the prompt, $P(C|\pi_i)$. So, for each category, only the item in the prompt (independent variable) will change across queries while the effect on the category probability is measured (dependent variable). After each prompting, the model input is flushed, guaranteeing that only the independent variable is manipulated for each trial for each category. This necessitates that the results be evaluated within category, since cross category results are not controlled. However, we also evaluate the test results across all categories for each model as a direct comparison to the results reported in the original paper.

\subsubsection{Individual Probability Correlation Test} 

For each population, we test for behavior consistent with typicality by evaluating the Pearson correlation between $P(C|\pi_i)$ and $r_i$ for all $i\in C$ and for all categories in the dataset. We hypothesize that, consistent with previous results, the probabilities output by the models will be positively correlated with typicality.

As predicted, all models show significant (p$<$0.05) probability/typicality correlation within nearly all categories consistent with typicality in humans in figure \ref{E1RQ1}. DistilBERT shows insignificant correlation with the categories tool and vegetable, while DistilGPT2 and RoBERTa base both have insignificant correlation with tool. More generally, the correlation between probability and typicality is strongly conditioned upon category for all models. The behavior shows strong differentiation between causal (CLM) and masked language models (MLMs). Among all MLMs the total correlation is markedly higher than for CLMs. Further, the categories for which each model most exhibits typicality behavior differs across MLMs and CLMs. 

The green bars in figure \ref{E1RQ1} represent the total correlation (across categories) obtained by evaluating only the base models and agree with past results \cite{misra2021language}. However, the total population correlation, shown in dark blue, suggests that the base model total correlation is an over estimation of the true total correlation.

\subsubsection{Population Uncertainty Correlation Test} 

We hypothesize that population uncertainty will be positively correlated with diminishing typicality. That is, as stimuli become less typical, the population will have decreasing agreement. Therefore, we test for correlation between normalized population standard deviation (as a measure of group uncertainty), $ \frac{\sigma(P(C|\pi_i))}{\mu(P(C|\pi_i))}$ and typicality rankings $r_i$. 

\begin{figure}[h]
\centering
\includegraphics[width=\columnwidth]{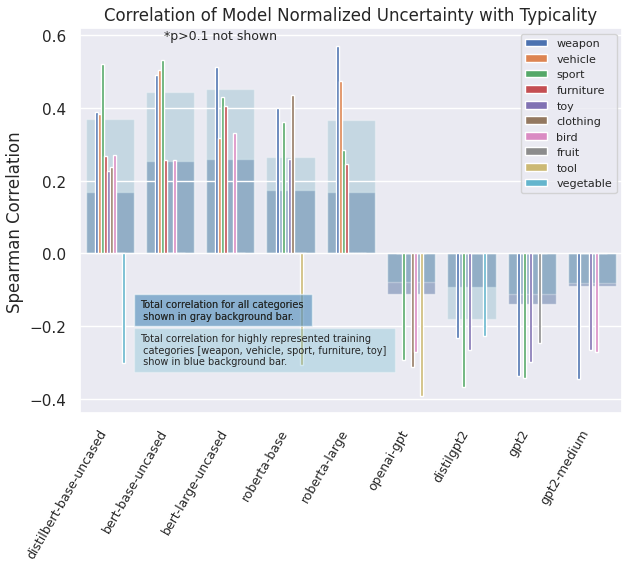} 
\caption{Within category and total Spearman correlation (p$<$0.08) is shown. The total Spearman correlation (p$<$0.01) for categories with an average item frequency $>$ 60000 in training data for Bert is also shown.}
\label{E1RQ4}
\end{figure}

In figure \ref{E1RQ4}, for masked language models, mean normalized population uncertainty has a significant positive correlation as typicality diminishes mediated by category. The models tend to become more uncertain as the items become less typical. Therefore, we believe that masked language models, like humans \cite{roschCognitive1975}, are more certain when inferencing about typical items. The categories which are most positively correlated with population certainty tend to be consistent with those which were most correlated with probability. 

Interestingly, the standard deviation of model probabilities was found to scale with the mean of the probability, giving the appearance of increased population agreement as probability declined. Therefore, mean normalization is used. Mean normalized uncertainty may be more meaningful than standard deviation alone for models which learn to output probabilities. 

In sharp distinction, all causal language models exhibit negative certainty/typicality correlation. We speculate that this may be due to differences in training data and modeling objective. i.e. it is not typical for humans to say extremely obvious things like "A sparrow is a bird." Therefore, a dropout trained conversational model may have high uncertainty regarding highly typical item/category pairings in completions. However, this hypothesis is not readily testable due to GPT-2 training data unavailability. Further, the categories among the CLMs which are most negatively correlated with population certainty do not seem to be the same categories as those which were most positively correlated with probability in figure \ref{E1RQ1}. This suggests that CLMs represent something all together different than MLMs in their population uncertainty.

\subsubsection{Confound Test} 

We considered that frequency of an item within the training data could act as a third variable and confound the results. To address this we evaluate the Pearson correlation of item frequency in the training data with typicality ranking. We hypothesized that item frequency would act as a confound at some level. We used the BERT family training data frequencies from \cite{zhou2022problems} to assess training data frequency correlations. 

We found no correlation between item typicality and frequency in the training corpus. Nor did we find a correlation between the normalized certainty and item frequency. There was a slight correlation (Spearman's $r$=-0.08 p$<$0.01) between probabilities output by BERT and item frequency. However, the effect size suggests that this is insignificant. 

\begin{figure}[h]
\centering
\includegraphics[width=\columnwidth]{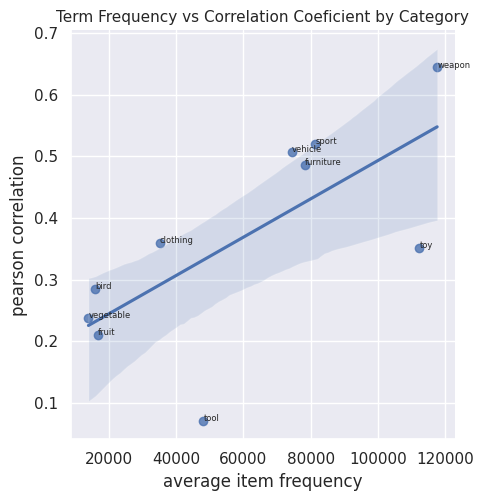} 
\caption{Emergence of within category behavior consistent with typicality in BERT is strongly predicted by within category item frequency in training data.}
\label{E1RQ3}
\end{figure}

We hypothesized that category "understanding" may be important for the emergence of typicality behavior and that mean within category item frequency in training data may predict category understanding. To test this, we performed a regression between within category Pearson correlation and mean within category item frequency in the training data for the BERT population.

In figure \ref{E1RQ3} we find that average item frequency within a category is highly correlated with the strength of typicality effects exhibited by the model within that category. The exception being the categories tool and toy. Further research may be necessary to determine why these categories do not fit the otherwise established trend. We suspect that this is the result of conflicts from the basic-level effect, that humans have a preferred level of categorization, which has a known relationship with typicality \cite{roschBasicObjectsNatural1976a}. Tool and toy may be outliers because they are not at the basic categorization level for many of the items listed in those categories.

If the anomalous categories are removed, the correlation between within category probability/typicality correlation and within category mean item frequency in training data is Pearson $r$=0.98 (p$<$0.01) and with tool and toy included Pearson $r$=0.7 (p$<$0.03). 

Another measure of concept "understanding" is the persistence of a concept through growing rates of dropout. As the dropout rate increases, more and more neurons are masked in the population members causing concepts with fewer constituent neurons to be ablated. So, we swept the dropout rate for the population from 0.1 to 0.8 and found that the categories which were highly represented in the training data tended to persist as dropout increased, while the categories with less training data representation tended to decorrelate at lower dropout rates.

We interpret these complimentary results to suggest that model "understanding" of a category may be driven by overall category representation in the training data and that, within categories which are well understood, models are likely to exhibit typicality effects. We find that this is the case for all tested model species as restricting the total Pearson correlation to the categories which are well represented in the BERT training data, a partial constituent of all other model's training, leads to significant increases in all model probability-typicality correlations in figure \ref{E1RQ1}. 

\subsubsection{Comments}

We find that language models strongly exhibit typicality effects both in individual model probabilities and in population uncertainty mediated by model "understanding" of category. The square of probability/typicality correlation in figure \ref{E1RQ1} shows that $10\%<r^2<25\%$ of the well represented category probability variances for all model populations, excluding GPT-1 which was trained on substantially less data, are accounted for by typicality effects. Strong typicality effects tend to emerge in categories with at least 80000 training examples per within category item.

\begin{table*}[!h]
\centering
\begin{tabular}{r|c|c|c|c}
    Model Species & Wilcoxon $P(S_x|\pi_x) > P(S_x)$ & $\frac{\mu(P(S_x|\pi_x))-\mu(P(S_x))}{\mu(P(S_x|\pi_y))-\mu(P(S_x))}$ & Pearson(AT-CT, PT-CT) $r$ & Structural Priming\\
    \toprule
    DistilBERT  & 0.42 (p$\approx$1) & --- & --- & None\\
    BERT Base  & 0.27 (p$\approx$1) & --- & --- & None \\
    BERT Large & 0.47 (p$\approx$1) & --- & --- & None \\
    
    \hline
    GPT & 0.14 (p$\approx$1) & --- & --- & None \\
    
    \hline
    DistilGPT-2  & 0.82 (p$\approx$0) & 0.93 $\pm$0.001 & 0.89 (p$<$0.01) & None \\
    GPT-2 & 0.96 (p$\approx$0) & 0.96 $\pm$0.001 & 0.93 (p$<$0.01) & None \\
    GPT-2 Medium & 0.99 (p$\approx$0) & 0.98 $\pm$0.001 & 0.94 (p$<$0.01) & None \\
    
    \hline
    RoBERTa Base & 0.99 (p$\approx$0) & 0.98 $\pm$0.001 & 0.70 (p$<$0.01) & Marginal \\
    RoBERTa Large & 0.99 (p$\approx$0) & 0.96 $\pm$0.001 & 0.67 (p$<$0.01) & Marginal  \\
    
\end{tabular}
\caption{Test results used to detect structural priming. From left to right: the first relates preference for priming; the second finds the percent of the preference magnitude not attributable to SP; the third measures the correlation between SP and an alternative.}
\label{tab:results}
\end{table*}

\section{Experiment 2: Structural Priming Effects in Language Models}

In \cite{sinclair2022structural} the authors investigated whether language models exhibit behavior consistent with the structural priming effect. We run a similar experiment using sentence data from their work. However, we use a dropout population, modify the experimental setup to control for unaddressed confounds, and perform a split-group cross validation. 

\subsubsection{Experimental Setup} 

To test for SP in language models we adopt 3 treatment conditions: the control (CT) is the probability of a sentence, $S_x$, without any priming $P(S_x)$; the primed treatment (PT) is the probability of that sentence when the language model is first prompted with a sentence, $\pi_x$, of similar structure $P(S_x|\pi_x)$; and the alternative treatment (AT) is the probability of $S_x$ when prompted with a sentence, $\pi_y$, of differing structure $P(S_x|\pi_y)$ but identical semantic meaning. Any effect AT has will not be analogous to SP. However, it is not a placebo as it may not be inert. Therefore, both AT and PT must be compared to CT for contextualization.  

We split 3000 examples into two groups and conduct all 3 treatments on all 50 population members per species. The results for the first group of 1500 are reported and the results for the second set of 1500 are used for cross validation. The cross validation showed all results repeated within $\pm$0.02 (p$<$0.01) of our reported results.

\subsubsection{Individual Probability Difference Test} 

For behavior consistent with SP to be present, the relationship PT$>$CT must tend to hold. To test this, we employ the Wilcoxon signed rank test, a non-parametric test appropriate for testing relative ranking of paired samples. 

In table \ref{tab:results} the results show that only GPT-2 and RoBERTa exhibit a preference for PT over the control. These models require subsequent testing as SP is one possible explanation for their preference, but an alternative hypothesis is that the models prefer being primed with anything at all. 

Preference for priming could be induced by the presence of WebText and OpenWebText in the training data of GPT-2 and RoBERTa families as these possess conversational data in which SP is more likely to be observed.

\subsubsection{Elimination of Alternative Hypotheses}

To eliminate a preference for priming regardless of structure as an alternative hypothesis, we find the 95\% confidence interval of $\frac{\mu(AT)-\mu(CT)}{\mu(PT)-\mu(CT)}$. This is the fraction of the probability change induced by PT which is not attributable to SP. Wilcoxon is not used in this case as it would result in the cancellation of the control group due to internal subtraction. It is possible that the effect magnitude will be similar but the individual samples not be correlated. Therefore, we also find the Pearson correlation between PT-CT and AT-CT. 

For all models the alternative treatment produced an average effect which was 96\% as large as the mean change due to SP. Further, the GPT-2 family showed strong correlation between AT-CT and PT-CT, suggesting these models do not prefer priming with a similar structure. However, the results in table \ref{tab:results} show that the RoBERTa family has a response to PT distinct from the AT response based on Pearson's $r$. 

\subsubsection{Comments}
In contrast to previous work we find little evidence for the presence of structural priming effects. The RoBERTa family of models exhibits a response that is distinct when primed with a sentence of similar structure to the target sentence. However, the preference magnitude is not differentiable from an alternative structure priming. No other models exhibit significant, distinct effects.

\section{Conclusions}

This paper addresses a current need in the study of cognitive behavior in neural models by introducing PopulationLM, a system built on MC dropout for the creation of efficient populations of neural models. This permits population based analysis of model behavior which may decrease the presence of atypical behaviors. In both experiments our population studies, when compared to the original experiments of other authors, show that conclusions drawn from single models tend to over estimate the presence of cognitive behaviors. Beyond robustness, populations permit the study of divergence or decorrelation as a function of dropout and characterization of population uncertainty or disagreement. 

We have conducted novel experiments using PopulationLM regarding the presence of typicality and structural priming in language models, being careful to isolate and analyze along independent variables and report effect sizes and significance. We find that typicality is consistently present while structural priming seems to not be, with both having predictable ties to behavior representation in training data.

PopulationLM may have further reaching applications 
beyond the study of cognitive behavior. Many papers have begun to systematically study prompt pattern effects \cite{white2023prompt}. These possess similar issues of robustness to cognitive studies and could benefit from study among a population. Further, it's possible that populations of models may serve as proxies for initial human behavior studies in the future. This could augment the ethical and financial efficacy of psycholinguistic research \cite{brysbaertHowManyParticipants2019}.

Test time augmentation \cite{gawlikowskiSurveyDeep2023} may be used to create local variations that perform similarly to dropout populations. However, the effects will decay with the length of the decoder context \cite{hahnTheoretical2020}. The longer the priming, the less effect each individual token, including the experimental prompt, will have. We intend to investigate the use of test time augmentation for closed source language model systematic population studies in future work. 

Finally, future work should investigate the
(1) surprising increase in CLM certainty with decreased typicality, (2) the use of mean normalization to characterize probabilistic model certainty, and (3) the presence and impact of other cognitive phenomena like basic level effects.

\bibliography{aaai24}

\begin{thebibliography}{36}
\providecommand{\natexlab}[1]{#1}

\bibitem[{Binz and Schulz(2023)}]{binz2023using}
Binz, M.; and Schulz, E. 2023.
\newblock Using cognitive psychology to understand GPT-3.
\newblock \emph{Proceedings of the National Academy of Sciences}, 120(6):
  e2218523120.

\bibitem[{Brysbaert(2019)}]{brysbaertHowManyParticipants2019}
Brysbaert, M. 2019.
\newblock How {Many} {Participants} {Do} {We} {Have} to {Include} in {Properly}
  {Powered} {Experiments}? {A} {Tutorial} of {Power} {Analysis} with
  {Reference} {Tables}.
\newblock \emph{Journal of Cognition}, 2(1): 16.

\bibitem[{Bubeck et~al.(2023)Bubeck, Chandrasekaran, Eldan, Gehrke, Horvitz,
  Kamar, Lee, Lee, Li, Lundberg et~al.}]{bubeck2023sparks}
Bubeck, S.; Chandrasekaran, V.; Eldan, R.; Gehrke, J.; Horvitz, E.; Kamar, E.;
  Lee, P.; Lee, Y.~T.; Li, Y.; Lundberg, S.; et~al. 2023.
\newblock Sparks of artificial general intelligence: Early experiments with
  gpt-4.
\newblock \emph{arXiv preprint arXiv:2303.12712}.

\bibitem[{Coda-Forno et~al.(2023)Coda-Forno, Witte, Jagadish, Binz, Akata, and
  Schulz}]{coda2023inducing}
Coda-Forno, J.; Witte, K.; Jagadish, A.~K.; Binz, M.; Akata, Z.; and Schulz, E.
  2023.
\newblock Inducing anxiety in large language models increases exploration and
  bias.
\newblock \emph{arXiv preprint arXiv:2304.11111}.

\bibitem[{Devlin et~al.(2018)Devlin, Chang, Lee, and
  Toutanova}]{devlin2018bert}
Devlin, J.; Chang, M.-W.; Lee, K.; and Toutanova, K. 2018.
\newblock Bert: Pre-training of deep bidirectional transformers for language
  understanding.
\newblock \emph{arXiv preprint arXiv:1810.04805}.

\bibitem[{Gal and Ghahramani(2016)}]{gal2016dropout}
Gal, Y.; and Ghahramani, Z. 2016.
\newblock Dropout as a bayesian approximation: Representing model uncertainty
  in deep learning.
\newblock In \emph{international conference on machine learning}, 1050--1059.
  PMLR.

\bibitem[{Gawlikowski et~al.(2023)Gawlikowski, Tassi, Ali, Lee, Humt, Feng,
  Kruspe, Triebel, Jung, Roscher, Shahzad, Yang, Bamler, and
  Zhu}]{gawlikowskiSurveyDeep2023}
Gawlikowski, J.; Tassi, C. R.~N.; Ali, M.; Lee, J.; Humt, M.; Feng, J.; Kruspe,
  A.; Triebel, R.; Jung, P.; Roscher, R.; Shahzad, M.; Yang, W.; Bamler, R.;
  and Zhu, X.~X. 2023.
\newblock A survey of uncertainty in deep neural networks.
\newblock \emph{Artificial Intelligence Review}.

\bibitem[{Goodman, Fanelli, and Ioannidis(2016)}]{goodman2016does}
Goodman, S.~N.; Fanelli, D.; and Ioannidis, J.~P. 2016.
\newblock What does research reproducibility mean?
\newblock \emph{Science translational medicine}, 8(341): 341ps12--341ps12.

\bibitem[{Hahn(2020)}]{hahnTheoretical2020}
Hahn, M. 2020.
\newblock Theoretical {Limitations} of {Self}-{Attention} in {Neural}
  {Sequence} {Models}.
\newblock \emph{Transactions of the Association for Computational Linguistics},
  8: 156--171.

\bibitem[{Jones and Steinhardt(2022)}]{jones2022capturing}
Jones, E.; and Steinhardt, J. 2022.
\newblock Capturing failures of large language models via human cognitive
  biases.
\newblock \emph{Advances in Neural Information Processing Systems}, 35:
  11785--11799.

\bibitem[{Kosinski(2023)}]{kosinski2023theory}
Kosinski, M. 2023.
\newblock Theory of mind may have spontaneously emerged in large language
  models.
\newblock \emph{arXiv preprint arXiv:2302.02083}.

\bibitem[{Lakshminarayanan, Pritzel, and
  Blundell(2017)}]{lakshminarayanan2017simple}
Lakshminarayanan, B.; Pritzel, A.; and Blundell, C. 2017.
\newblock Simple and scalable predictive uncertainty estimation using deep
  ensembles.
\newblock \emph{Advances in neural information processing systems}, 30.

\bibitem[{Lamprinidis(2023)}]{lamprinidis2023llm}
Lamprinidis, S. 2023.
\newblock LLM Cognitive Judgements Differ From Human.
\newblock \emph{arXiv preprint arXiv:2307.11787}.

\bibitem[{Liu et~al.(2019)Liu, Ott, Goyal, Du, Joshi, Chen, Levy, Lewis,
  Zettlemoyer, and Stoyanov}]{liu2019roberta}
Liu, Y.; Ott, M.; Goyal, N.; Du, J.; Joshi, M.; Chen, D.; Levy, O.; Lewis, M.;
  Zettlemoyer, L.; and Stoyanov, V. 2019.
\newblock Roberta: A robustly optimized bert pretraining approach.
\newblock \emph{arXiv preprint arXiv:1907.11692}.

\bibitem[{Lyzhov et~al.(2020)Lyzhov, Molchanova, Ashukha, Molchanov, and
  Vetrov}]{lyzhov2020greedy}
Lyzhov, A.; Molchanova, Y.; Ashukha, A.; Molchanov, D.; and Vetrov, D. 2020.
\newblock Greedy policy search: A simple baseline for learnable test-time
  augmentation.
\newblock In \emph{Conference on Uncertainty in Artificial Intelligence},
  1308--1317. PMLR.

\bibitem[{McCoy, Pavlick, and Linzen(2019)}]{mccoy2019right}
McCoy, R.~T.; Pavlick, E.; and Linzen, T. 2019.
\newblock Right for the wrong reasons: Diagnosing syntactic heuristics in
  natural language inference.
\newblock \emph{arXiv preprint arXiv:1902.01007}.

\bibitem[{Misra, Ettinger, and Rayz(2021)}]{misra2021language}
Misra, K.; Ettinger, A.; and Rayz, J.~T. 2021.
\newblock Do language models learn typicality judgments from text?
\newblock \emph{arXiv preprint arXiv:2105.02987}.

\bibitem[{Pickering and Ferreira(2008)}]{pickering2008structural}
Pickering, M.~J.; and Ferreira, V.~S. 2008.
\newblock Structural priming: a critical review.
\newblock \emph{Psychological bulletin}, 134(3): 427.

\bibitem[{Radford et~al.(2018)Radford, Narasimhan, Salimans, Sutskever
  et~al.}]{radford2018improving}
Radford, A.; Narasimhan, K.; Salimans, T.; Sutskever, I.; et~al. 2018.
\newblock Improving language understanding by generative pre-training.
\newblock \emph{OpenAI blog}.

\bibitem[{Radford et~al.(2019)Radford, Wu, Child, Luan, Amodei, Sutskever
  et~al.}]{radford2019language}
Radford, A.; Wu, J.; Child, R.; Luan, D.; Amodei, D.; Sutskever, I.; et~al.
  2019.
\newblock Language models are unsupervised multitask learners.
\newblock \emph{OpenAI blog}, 1(8): 9.

\bibitem[{Rosch(1975)}]{roschCognitive1975}
Rosch, E. 1975.
\newblock Cognitive representations of semantic categories.
\newblock \emph{Journal of Experimental Psychology: General}, 104(3): 192--233.

\bibitem[{Rosch et~al.(1976)Rosch, Mervis, Gray, Johnson, and
  Boyes-Braem}]{roschBasicObjectsNatural1976a}
Rosch, E.; Mervis, C.~B.; Gray, W.~D.; Johnson, D.~M.; and Boyes-Braem, P.
  1976.
\newblock Basic objects in natural categories.
\newblock \emph{Cognitive Psychology}, 8(3): 382--439.

\bibitem[{Sanh et~al.(2019)Sanh, Debut, Chaumond, and
  Wolf}]{sanh2019distilbert}
Sanh, V.; Debut, L.; Chaumond, J.; and Wolf, T. 2019.
\newblock DistilBERT, a distilled version of BERT: smaller, faster, cheaper and
  lighter.
\newblock \emph{arXiv preprint arXiv:1910.01108}.

\bibitem[{Sap et~al.(2022)Sap, LeBras, Fried, and Choi}]{sap2022neural}
Sap, M.; LeBras, R.; Fried, D.; and Choi, Y. 2022.
\newblock Neural theory-of-mind? on the limits of social intelligence in large
  lms.
\newblock \emph{arXiv preprint arXiv:2210.13312}.

\bibitem[{Sharir, Peleg, and Shoham(2020)}]{sharir2020cost}
Sharir, O.; Peleg, B.; and Shoham, Y. 2020.
\newblock The cost of training nlp models: A concise overview.
\newblock \emph{arXiv preprint arXiv:2004.08900}.

\bibitem[{Shelmanov et~al.(2021)Shelmanov, Tsymbalov, Puzyrev, Fedyanin,
  Panchenko, and Panov}]{shelmanovHowCertainYour2021}
Shelmanov, A.; Tsymbalov, E.; Puzyrev, D.; Fedyanin, K.; Panchenko, A.; and
  Panov, M. 2021.
\newblock How {Certain} is {Your} {Transformer}?
\newblock In \emph{Proceedings of the 16th {Conference} of the {European}
  {Chapter} of the {Association} for {Computational} {Linguistics}: {Main}
  {Volume}}, 1833--1840. Online: Association for Computational Linguistics.

\bibitem[{Sinclair et~al.(2022)Sinclair, Jumelet, Zuidema, and
  Fern{\'a}ndez}]{sinclair2022structural}
Sinclair, A.; Jumelet, J.; Zuidema, W.; and Fern{\'a}ndez, R. 2022.
\newblock Structural persistence in language models: Priming as a window into
  abstract language representations.
\newblock \emph{Transactions of the Association for Computational Linguistics},
  10: 1031--1050.

\bibitem[{Subramanian, Chitlangia, and
  Baths(2022)}]{subramanian2022reinforcement}
Subramanian, A.; Chitlangia, S.; and Baths, V. 2022.
\newblock Reinforcement learning and its connections with neuroscience and
  psychology.
\newblock \emph{Neural Networks}, 145: 271--287.

\bibitem[{Suri et~al.(2023)Suri, Slater, Ziaee, and Nguyen}]{suri2023large}
Suri, G.; Slater, L.~R.; Ziaee, A.; and Nguyen, M. 2023.
\newblock Do Large Language Models Show Decision Heuristics Similar to Humans?
  A Case Study Using GPT-3.5.
\newblock \emph{arXiv preprint arXiv:2305.04400}.

\bibitem[{Trott et~al.(2023)Trott, Jones, Chang, Michaelov, and
  Bergen}]{trott2023large}
Trott, S.; Jones, C.; Chang, T.; Michaelov, J.; and Bergen, B. 2023.
\newblock Do Large Language Models know what humans know?
\newblock \emph{Cognitive Science}, 47(7): e13309.

\bibitem[{Ullman(2023)}]{ullman2023large}
Ullman, T. 2023.
\newblock Large language models fail on trivial alterations to theory-of-mind
  tasks.
\newblock \emph{arXiv preprint arXiv:2302.08399}.

\bibitem[{Vaswani et~al.(2017)Vaswani, Shazeer, Parmar, Uszkoreit, Jones,
  Gomez, Kaiser, and Polosukhin}]{vaswani2017attention}
Vaswani, A.; Shazeer, N.; Parmar, N.; Uszkoreit, J.; Jones, L.; Gomez, A.~N.;
  Kaiser, {\L}.; and Polosukhin, I. 2017.
\newblock Attention is all you need.
\newblock \emph{Advances in neural information processing systems}, 30.

\bibitem[{Vazhentsev et~al.(2022)Vazhentsev, Kuzmin, Shelmanov, Tsvigun,
  Tsymbalov, Fedyanin, Panov, Panchenko, Gusev, Burtsev, Avetisian, and
  Zhukov}]{vazhentsevUncertainty2022}
Vazhentsev, A.; Kuzmin, G.; Shelmanov, A.; Tsvigun, A.; Tsymbalov, E.;
  Fedyanin, K.; Panov, M.; Panchenko, A.; Gusev, G.; Burtsev, M.; Avetisian,
  M.; and Zhukov, L. 2022.
\newblock Uncertainty {Estimation} of {Transformer} {Predictions} for
  {Misclassification} {Detection}.
\newblock In \emph{Proceedings of the 60th {Annual} {Meeting} of the
  {Association} for {Computational} {Linguistics} ({Volume} 1: {Long}
  {Papers})}, 8237--8252. Dublin, Ireland: Association for Computational
  Linguistics.

\bibitem[{Wei et~al.(2022)Wei, Tay, Bommasani, Raffel, Zoph, Borgeaud,
  Yogatama, Bosma, Zhou, Metzler et~al.}]{wei2022emergent}
Wei, J.; Tay, Y.; Bommasani, R.; Raffel, C.; Zoph, B.; Borgeaud, S.; Yogatama,
  D.; Bosma, M.; Zhou, D.; Metzler, D.; et~al. 2022.
\newblock Emergent abilities of large language models.
\newblock \emph{arXiv preprint arXiv:2206.07682}.

\bibitem[{White et~al.(2023)White, Fu, Hays, Sandborn, Olea, Gilbert, Elnashar,
  Spencer-Smith, and Schmidt}]{white2023prompt}
White, J.; Fu, Q.; Hays, S.; Sandborn, M.; Olea, C.; Gilbert, H.; Elnashar, A.;
  Spencer-Smith, J.; and Schmidt, D.~C. 2023.
\newblock A prompt pattern catalog to enhance prompt engineering with chatgpt.
\newblock \emph{arXiv preprint arXiv:2302.11382}.

\bibitem[{Zhou et~al.(2022)Zhou, Ethayarajh, Card, and
  Jurafsky}]{zhou2022problems}
Zhou, K.; Ethayarajh, K.; Card, D.; and Jurafsky, D. 2022.
\newblock Problems with cosine as a measure of embedding similarity for high
  frequency words.
\newblock \emph{arXiv preprint arXiv:2205.05092}.

\end{thebibliography}

\end{document}